\pdfoutput=1

\documentclass[11pt]{article}

\usepackage{acl}
\usepackage{times}
\usepackage{latexsym}

\usepackage{graphicx}
\usepackage{booktabs} 
\usepackage{multirow}

\usepackage[T1]{fontenc}

\usepackage[utf8]{inputenc}

\usepackage{microtype}

\usepackage{inconsolata}

\title{PeriodicLoRA: Breaking the Low-Rank Bottleneck in LoRA Optimization}

\author{Xiangdi Meng$^{1, 2}$, Damai Dai$^{1}$, Weiyao Luo$^{1, 2}$, Zhe Yang$^{1}$, Shaoxiang Wu$^{1, 2}$\\ 
\textbf{Xiaochen Wang}$^{1, 2}$, \textbf{Peiyi Wang}$^{1}$, \textbf{Qingxiu Dong}$^{1}$, \textbf{Liang Chen}$^{1}$, \textbf{Zhifang Sui}$^{1}$\footnotemark[1] \\
$^1$National Key Laboratory for Multimedia Information Processing, \\
School of Computer Science, Peking University \\ 
$^2$School of Software \& Microelectronics, Peking University\\
\texttt{xd\_meng@outlook.com}\\ 
\texttt{\{daidamai, szf\}@pku.edu.cn}\\
}

\begin{document}
\maketitle
\begin{abstract}
Supervised fine-tuning is the most common method to adapt large language models (LLMs) to downstream tasks, but full fine-tuning LLMs requires massive computational resources.
Recently, parameter-efficient fine-tuning (PEFT) methods have been widely studied due to its cost-effectiveness. 
LoRA is one of the most widely used methods, which assumes that the optimization process is essentially low-dimensional. 
Although LoRA fine-tuning is effective, there is still a performance gap compared to full fine-tuning, since its weight update is limited to low-rank matrices.
In order to break the low-rank bottleneck in LoRA Optimization, we propose PeriodicLoRA (PLoRA), which accumulates low-rank update matrices multiple times to achieve a higher update rank. 
PLoRA has multiple training stages. 
During each stage, we still update only the LoRA weights.
However, at the end of each stage, we unload the LoRA weights into the backbone parameters and then reinitialize the LoRA states. 
Experimental results show that PLoRA has stronger learning ability, approximately 1.8 times that of LoRA's learning ability at most, but it does not increase memory usage.
Further, we introduce a momentum-based unloading strategy for PLoRA to mitigate the training instability. 

\end{abstract}
\footnotetext[1]{Corresponding author.}
\section{Introduction}
Large language models are becoming increasingly proficient in natural language processing \citep{openai2023gpt4,chen2024chatgpts}, leading to a growing demand for their application in various downstream tasks. 
Supervised fine-tuning is the most widely used adaptation method, but fully fine-tuning large models is challenging due to its heavy cost on computational resources and memory usage. 

To address this issue, several parameter-efficient fine-tuning (PEFT) methods have been developed. 
PEFT methods can achieve considerable fine-tuning performance, but require significantly fewer computational resources than full fine-tuning. 
LoRA is currently the most widely used PEFT method. 
It allocates additional low-rank matrices for the backbone model, and only optimizes the low-rank matrices during training. 
With LoRA, the memory usage required for fine-tuning a model can be significantly reduced, enabling fine-tuning studies on large language models within limited computational resources.

Although LoRA can achieve considerable fine-tuning performance, it still lags behind full fine-tuning. 
The largest bottleneck lies in that the LoRA update matrix $\Delta W$ usually features a low rank. 
When $\Delta W$ has too low a rank, the learning capability of LoRA will go weak compared to the full fine-tuning method, which can produce full-rank update matrix. 
If we can increase the rank of the update matrix $\Delta W$, the learning capability of LoRA-like methods will gradually approach to that of full fine-tuning. 
However, directly increasing the LoRA rank will also increase memory overhead. 

In this paper, we propose a new PEFT method, PeriodicLoRA (PLoRA), to increase the rank of the update matrix $\Delta W$ while maintaining constant memory usage. 
The key idea of PLoRA is to periodically unload the LoRA weights trained on mini-batches into the backbone parameters, and the accumulation of multiple low-rank update matrices can produce a higher-rank update matrix.
To be specific, PLoRA has multiple training stages. 
During each stage, we still update only the LoRA weights.
However, at the end of each stage, we unload the LoRA weights into the backbone parameters and then reinitialize the LoRA states, including the LoRA weights, corresponding optimizer state, and learning rate scheduler state. 
Theoretically, PLoRA allows the LoRA method to break through the limitation of the low-rank update matrix and to approach the effect of full fine-tuning.

\begin{figure*}[!t]
    \centering
    \includegraphics[width=\textwidth]{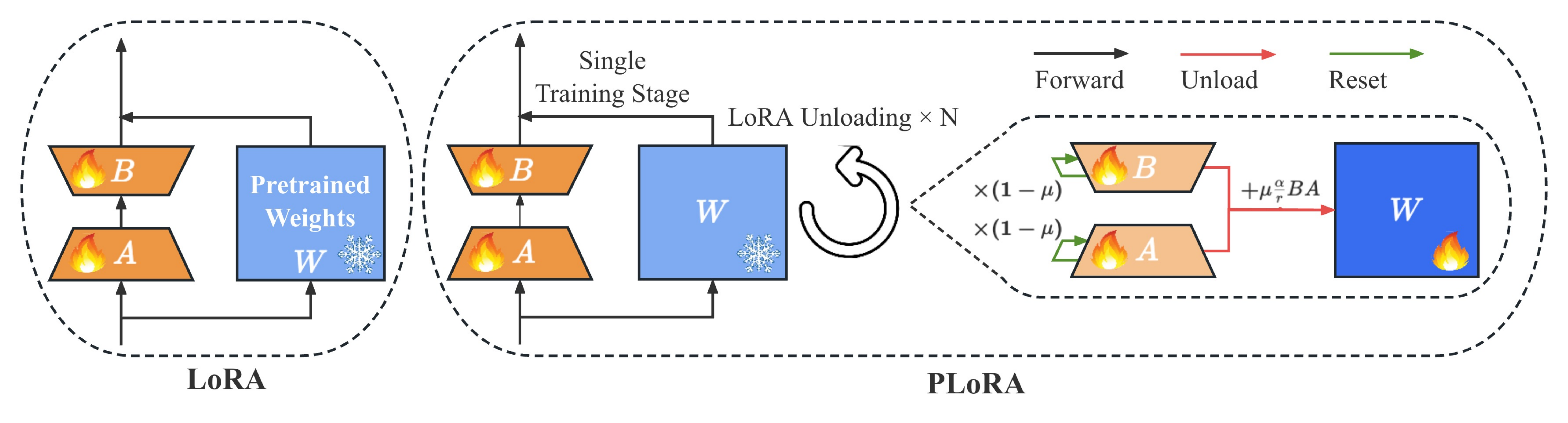}
    \caption{Compared to LoRA, the proposed PLoRA structure is outlined. In LoRA training, only matrices $A$ and $B$ are updated while the model weights are frozen (signified by blue). The trained matrices $A$ and $B$ are then used to update the model weights after completing all training. While in the PLoRA approach, after each mini-batch training, the weights of matrices $A$ and $B$ in LoRA are transferred to update the model weights and reset themselves before continuing training. This cycle repeats N times within one epoch. In the figure, light orange represents learning ability on a mini-batch basis, while orange represents cumulative learning ability up to the current moment.
    }
    \label{fig:PLoRA}
\end{figure*}

In order to validate the effectiveness of PLoRA, we perform instruction tuning for LLaMA 7B~\citep{touvron2023llama} in different PEFT settings and evaluate the performance on multi-subject multiple-choice, math reasoning, and language understanding and reasoning tasks.
Experimental results show that PLoRA consistently outperforms LoRA with the same rank, but does not introduce any memory overhead.
In order to demonstrate our method more clearly and explain its capabilities, we discuss the ablation studies on PLoRA settings, and most importantly, the improvement of learning ability after applying PLoRA in different tasks.
Our contributions are summarized as follows:
\begin{itemize}
    \item We introduce PLoRA for parameter-efficient fine-tuning, which breaks through the low-rank bottleneck in LoRA fine-tuning. 
    \item We validate the effectiveness of PLoRA in instruction tuning, and show that it consistently outperforms the naive LoRA method.
    \item We conduct elaborate analysis on the training process of PLoRA, and reveal that PLoRA has stronger learning capability than naive LoRA.
    \item We make our full results in tuning hyper-parameters public to provide a reference for selecting a proper PLoRA setting. 
\end{itemize}

\section{Related Work}
\subsection{Supervised Fine-Tuning}
Large Language Models (LLMs) are known for their vast knowledge bases and text generation capabilities, yet they have traditionally exhibited subpar instruction-following proficiency. Past fine-tuning methods mainly focus on one special task \cite{GLUE} (i.e., single-task specialization), which means they can't handle various human instructions (i.e., multi-task generalization). Recent strides in instruction tuning have sought to extend the versatility of LLMs to better handle multi-task generalization. This instruction-tuning paradigm leverages datasets containing diverse instruction-output pairings, such as Self-instruct \cite{self-instruct}, UnnaturalInstruction \cite{unnatural-instruction}, Super-NaturalInstructions \cite{superNI}, and FLAN v2 \cite{FLAN}, thereby enhancing the models' comprehension of varied instructions.  Many recent instruction-tuned models include InstructGPT \cite{InstructGPT}, Guanaco \cite{Guanaco}, Vicuna \cite{vicuna}, and they have achieved a great performance in understanding general knowledge across a wide variety of fields from OpenLLM Leaderboard. Origin Lora is not good at multi task training, and it was also difficult to adjust the parameter rank. Therefore, Multilora \cite{multilora} reduces the dominance of top singular vectors observed in LoRA and scales LoRA modules horizontally to adapt multi task. Our work is simpler and more efficient by contrast.

\subsection{Parameter-Efficient Fine-Tuning}
Parameter-efficient tuning (PEFT) is a popular methods for training models which only optimize a small portion of parameters and keep the main model frozen for adaptation. Some PEFT methods would insert additional neural modules or parameters to the backbone model, such as Adapter \cite{peft}. For example, Prefix tuning \cite{prefixtuning} adds prefix parameters to the hidden states in every layer of the model. Prompt tuning \cite{prompt_tuning} uses template to reconstruct prompt, and only updates parameters related to prompt understanding. Besides, another way attempts to specify particular parameters to be trainable \cite{bitfit}.

\subsection{LoRA and Its Variants}
LoRA proposes a low-rank up and down projection transformation without any non-linearity applied in parallel to key and value attention matrices. The main benefit of LoRA is the adapter module, which can be integrated into the original weight matrices of the model, leading to a very efficient inference time. 

Although training and inference efficiency have greatly improved, LoRA's flexibility in choosing the ideal rank r remains a limitation. LoRA rank r accepts discrete values, the changes to which directly affect the model structures; in contrast, continuous hyperparameters, like learning rate, can be adjusted adaptively online during the training process. For various downstream activities and backbone models, there may be differences in the ideal rank selection. Regular ways of searching can squander computational resources and training time.

Several solutions have been proposed in recent years to enable the flexible tuning of LoRA rank. Dylora \cite{dylora} propose a new methodology for training low rank modules for multiple ranks simultaneously rather than training a single-rank adapter at a time (without changing the training budget). AdaLoRA \cite{adalora} introduces an additional regularizer to ensure that the lower and upper projection matrices strictly adhere to the definition of singular value decomposition (SVD), with each matrix being orthogonal. SparseLORA\cite{sparse-lora} eliminate this computational overhead and instead selectively filter low-rank components by controlling the intermediate diagonal matrix. LoraHub \cite{lorahub} proposes a strategy to automatically construct LoRA modules for a model in fine-tuning with diverse given tasks. 

\section{Method: PeriodicLoRA}

Firstly, Section \ref{method: lora} introduces traditional LoRA and its details as the background, and points out the low-rank bottleneck problem of traditional LoRA. Then, Section \ref{method: plora} introduces our PLoRA method as an effective solution to this problem.
\subsection{Preliminary: Naive LoRA Method}
\label{method: lora}
\paragraph{Motivation to utilize LoRA.} As the number of parameters in language models continues to increase, full-parameter fine-tuning of these models has become computationally expensive and requires GPUs with substantial memory capacity, which is impractical for researchers with limited computational resources. To facilitate efficient fine-tuning of neural networks in low-resource settings, while simultaneously avoiding the increased inference latency associated with traditional adapters, \citet{lora} proposes a low-rank adapter methodology for dense parameter matrices. A neural network typically comprises numerous dense parameter matrices. For a pre-trained parameter matrix $W_0 \in R^{d \times k}$, after full-parameter fine-tuning on a specific domain task, the new parameter matrix is \begin{equation} W_{ft} = W_0 + \Delta W, \end{equation} where $\Delta W$ represents the update containing domain knowledge. Empirically, both $W_0$ and $\Delta W$ tend to be full-rank.
\paragraph{Details and principles of naive LoRA.} LoRA approximates the update $\Delta W$ by decomposing it into the product of two low-rank matrices, constraining the update to a low-rank space and making the fine-tuning process more efficient. Through this approximate low-rank decomposition, we have \begin{equation} \Delta W \approx BA, \end{equation} where $B \in R^{d \times r}, A \in R^{r \times k}$, and the rank $r \ll max(d,k)$. In the forward propagation of LoRA, for an input representation $x$, the output after passing through the parameter matrix $W_0$ is \begin{equation} h = W_0x + BAx.\end{equation} To ensure that introducing LoRA at the initial stage does not impact the computation results of the model’s forward propagation, it is crucial to ensure that $BAx = 0$. To achieve this, LoRA initializes $A$ as a random Gaussian matrix and $B$ as a zero matrix. During training, $W_0$ is frozen, and $B$ and $A$ are treated as trainable parameters. Since $r$ is significantly smaller than $k$ or $d$, the parameter number of $B$ and $A$ is much less than that of $W_0$. Fine-tuning only a small number of parameters can substantially reduce the GPU memory requirement. Upon completion of training, the parameter matrices $A$ and $B$ are merged into $W_0$ to form the final parameter matrix \begin{equation} W_{ft} = W_0 + BA.\end{equation} It is noteworthy that the final update is $BA$, which is constrained within a low-rank space. 
\paragraph{Low-rank bottleneck of LoRA} Compressing the update $\Delta W$ into a low-rank space is not lossless. Although \citet{aghajanyan2020intrinsic,lora} claims that for some fine-tuning adaptations to specific tasks, the required update can be accomplished within low-rank space, and LoRA has indeed demonstrated similar performance to full-parameter fine-tuning on some simple text classification tasks, we still observe a considerable performance gap between LoRA and full-parameter fine-tuning on more complex tasks. For intricate tasks like GSM8K\citep{cobbe2021training}, the parameter updates necessary for model adaptation cannot be fully represented within the low-rank space, and Forced compression into this constrained space results in performance degradation. 

\subsection{PeriodicLoRA}
\label{method: plora}
To overcome the inherent low-rank bottleneck in the traditional LoRA approach, we introduce the PeriodicLoRA (PLoRA) method, as shown in Figure \ref{fig:PLoRA}. We divide the fine-tuning on downstream tasks into $T$ distinct stages, denoted as $1, 2, ..., T$, each containing a fixed number of steps. Each stage can be viewed as a full process of traditional LoRA, which comprises of initializing $B$ and $A$ matrices, fine-tuning them, and merging them into parameter matrix $W$.
Initially, we have the parameter matrix $W_0$; after $t-1$ stages, the parameter matrix is updated to $W_{t-1}$. At stage $t$, we initialize $B_t$ as a zero matrix and $A_t$ as a Gaussian matrix; after fine-tuning steps of stage $t$, trainable parameter $B_t, A_t$ tend to be convergent, and then we merge the matrix product $B_tA_t$ into the parameter matrix $W_{t-1}$, resulting in the updated parameter matrix \begin{equation}W_t = W_{t-1} + B_tA_t.\end{equation} In the subsequent stage, we repeat this process, initializing $B$ and $A$ matrices, tuning them on training data, and eventually merging them into the parameter matrix $W$. After $T$ stages, the final parameter matrix is \begin{equation} W_T = W_0 + B_1A_1 + B_2A_2 + ... + B_TA_T.\end{equation} Theoretically, the rank of the final update by our method is $T \times r$, and by increasing the number of stages $T$, we can break through the low-rank constraints of traditional LoRA.

\section{Experiments}
\subsection{Datasets}

\begin{table*}[!t]
\centering
\resizebox{0.95\textwidth}{!}{
\begin{tabular}{@{}l|r|c|ccccc|cc@{}}
\toprule
\textbf{Method} & \textbf{Trainable Params} & \textbf{GSM8K} & \textbf{Hum.} & \textbf{STEM} & \textbf{Social Sciences} & \textbf{Other} & \textbf{Average} &\textbf{ARC-e}& \textbf{ARC-c}\\ \midrule
Llama-7b & - & 0 & 24.9 & 22.4 & 23.1 & 23.4 & 23.6 &64.8 &34.9 \\
Full Fine-tuning & 6741.0M & 28.0 & 38.8 & 34.4 & 45.8 & 47.8 & 41.3 &72.8 &37.9 \\ \midrule
LoRA (rank = 1) & 2.5M & 16.3 & 34.6 & 33.7 & 41.0 & 44.5 & 38.0 &68.4 &35.9 \\
PLoRA (rank = 1) & 2.5M & 17.8 & 35.7 & 33.9 & 42.1 & 45.2 & 38.8 &70.6 &36.9\\
LoRA (rank = 8) & 20.0M & 20.6 & 37.6 & 33.9 & 47.7 & 47.8 & 41.2 &71.6 &37.8 \\
PLoRA (rank = 8) & 20.0M & 24.0 & 39.0 & 33.8 & 45.8 & 48.2 & 41.4 &71.7 &37.5 \\ \bottomrule
\end{tabular}}
\caption{Results of zero-shot evaluations for PLoRA, LoRA, and full fine-tuning. The MMLU task consists of 57 subtasks, which are further divided into 4 major categories: Humanities, STEM, Social Sciences, and Other. The Average in the table represents the average indicator of MMLU across all tasks. It should be noted that for the sake of fairness in experimental comparison here, the prompt format from the training set is uniformly used in generating evaluation results.}
\label{table:results}
\end{table*}

In order to make a more intuitive comparison with the currently commonly used benchmark for evaluating large models, and to avoid using large-scale datasets that would lengthen the training and evaluation cycle, we organize a multitask dataset consisting of a total of 60,542 entries. 
29,683 entries of this dataset are from Orca \citep{mukherjee2023orca} from Tülu \citep{wang2023far}. 

The Orca dataset is a dataset that samples user queries from FLAN and collects ChatGPT \citep{InstructGPT} responses. It has a positive effect on improving Massive Multitask Language Understanding (MMLU) \citep{hendrycks2020measuring}. However, due to the large size of the Orca dataset, we choose a subset of the Orca dataset sampled from Tülu as a replacement. Tülu is a mixed dataset that collects high-quality instruction fine-tuning datasets in proportion.
And we also include 20,016 entries from code-alpaca in the Tülu dataset.

In order to enhance the ability of the model to learn complex tasks, we also transform the training sets of GSM8K and ARC \citep{clark2018think} into dialogue format. There are 7,473 and 3,370 entries respectively included in our training set.
The dataset predominantly features human-to-GPT conversations, with the Orca from Tülu and code-alpaca from Tülu subsets being particularly challenging due to their complexity and the existence of associated research literature.

\subsection{Strategic Organization of Experimental Datasets}
The strategic organization of our experimental datasets is driven by the need to demonstrate the effectiveness of our proposed PLoRA method on challenging benchmarks. The inclusion of the Tülu datasets, which are well-documented in the literature, ensures that our method is tested against established benchmarks and that our findings are comparable with existing research. The selection of these datasets is not merely for the sake of achieving higher scores on the benchmarks but is also intended to showcase the potential of PLoRA to surpass the rank limitations imposed by LoRA, thereby enhancing the learning capabilities of our models.

It is important to note that our primary focus is on exploring the potential of PLoRA to break through the rank constraints set by LoRA and to achieve stronger learning capabilities. The organization of our datasets is primarily aimed at ensuring that we can gain benefits on the more challenging benchmarks without significantly impacting our iteration speed. While the use of superior data and appropriate data balancing can contribute to better experimental outcomes, these are not the central concerns of our study. Our emphasis is on the methodological advancement that PLoRA represents, and how it can lead to improved performance on complex tasks, as evidenced by our evaluations on the GSM8K \citep{cobbe2021training} , MMLU \citep{hendrycks2020measuring} and ARC \cite{clark2018think} datasets.

\subsection{Experimental Setup}
We select the LLaMA-7B \citep{touvron2023llama} as the base LLM and fine-tune it using PLoRA, LoRA and full fine-tuning respectively. We apply low-rank matrices to all attention layers and MLP linear layers, including the $W_q$, $W_k$, $W_v$, $W_o$, $W_{gate}$, $W_{up}$, $W_{down}$, implying that these layers undergo an unload process. The reason for this part will be further explained in Section~\ref{sec:LoRAs_Settings}. We implement PLoRA and LoRA with low-rank matrix values of 1 and 8. 
In terms of the optimizer configuration, we opt for the AdamW optimizer \citep{loshchilov2018fixing} with a constant learning rate of 1e-4. Our training framework is based on FastChat~\citep{zheng2023judging}. 
Using a constant learning rate facilitates experimental settings and ensures that training can continue on previous results when extending the training time.
Additionally, the adam $\beta_2$ is set to 0.99. The global batch size is set to 24. Our experiments are conducted using the HuggingFace's transformers library \citep{wolf2020transformers} on 8 A100 GPUs.
We use lm-evaluation-harness~\citep{eval-harness} to evaluate the results. 

\subsection{Results}
We first compare the zero-shot and few-shot capabilities of LLaMA-7B, followed by a comparison of the fine-tuning results using PLoRA, LoRA, and full fine-tuning methods. For PLoRA and LoRA, we conduct experiments with ranks of 1 and 8 to further validate the effectiveness of our method. The detailed experimental results are shown in Table~\ref{table:results}, with all results obtained after training for one epoch. 
Due to limitations in the dataset and model, we mainly evaluate benchmarks with more positive impacts on multitask datasets. Unfortunately, although code-alpaca's entries account for one-third of the dataset, our models still perform at around 1\% in terms of pass@20 in HumanEval tests. Therefore, we did not showcase the evaluation results of HumanEval.

We can observe that, compared to LoRA, our proposed PLoRA achieved performance improvements on both GSM8K and MMLU datasets, especially when the rank is set to 8. On the GSM8K dataset, our method achieve a remarkable 20\% improvement over the traditional LoRA. Surprisingly, the PLoRA method even outperforms full fine-tuning in some scenarios. On a more comprehensive level, while the MMLU average results of LoRA at rank 8 are inferior to full fine-tuning, the PLoRA method can achieve comparable or better performance with resource conservation.

\section{Discussion}
To fully analyze and unlock the potential of PLoRA, we discuss the main challenges and optimal parameter settings in the implementation of PLoRA.

\subsection{Unloading Points Selection}

\label{sec:}
The determination of unloading points---the amount of data used for unloading, is one of the most fundamental problem in applying PLoRA.
Using less data for unloading allows for additional unloads per training epoch, potentially increasing the achievable performance. However, this presents a trade-off, as insufficient data for unloading can lead to  insufficient training and error accumulation, ultimately resulting in training collapse.

\begin{table}[t]
    \centering
    \resizebox{0.35\textwidth}{!}{
    \begin{tabular}{@{}c|cc@{}}
    \toprule
        \textbf{Method} &\textbf{Unloading Data} & \textbf{GSM8K} \\
        \midrule
        \multirow{3}{*}{PLoRA} & 4.8k & 1.7 \\
        ~ & 9.6k & 9.9 \\
        ~ & 19.2k & 9.2 \\
        \midrule
        \multirow{1}{*}{LoRA} & 60.5k & 3.5 \\
        \bottomrule
    \end{tabular}}
    \caption{Fixed the rank of 1, after fine-tuning only $W_q$ and $W_v$ for one epoch, the performance of the model on the GSM8K test set with 0-shot setting.}
    \label{table:unloading}
\end{table}


\begin{figure}[t]
  \begin{center}
    \includegraphics[width=0.42\textwidth]{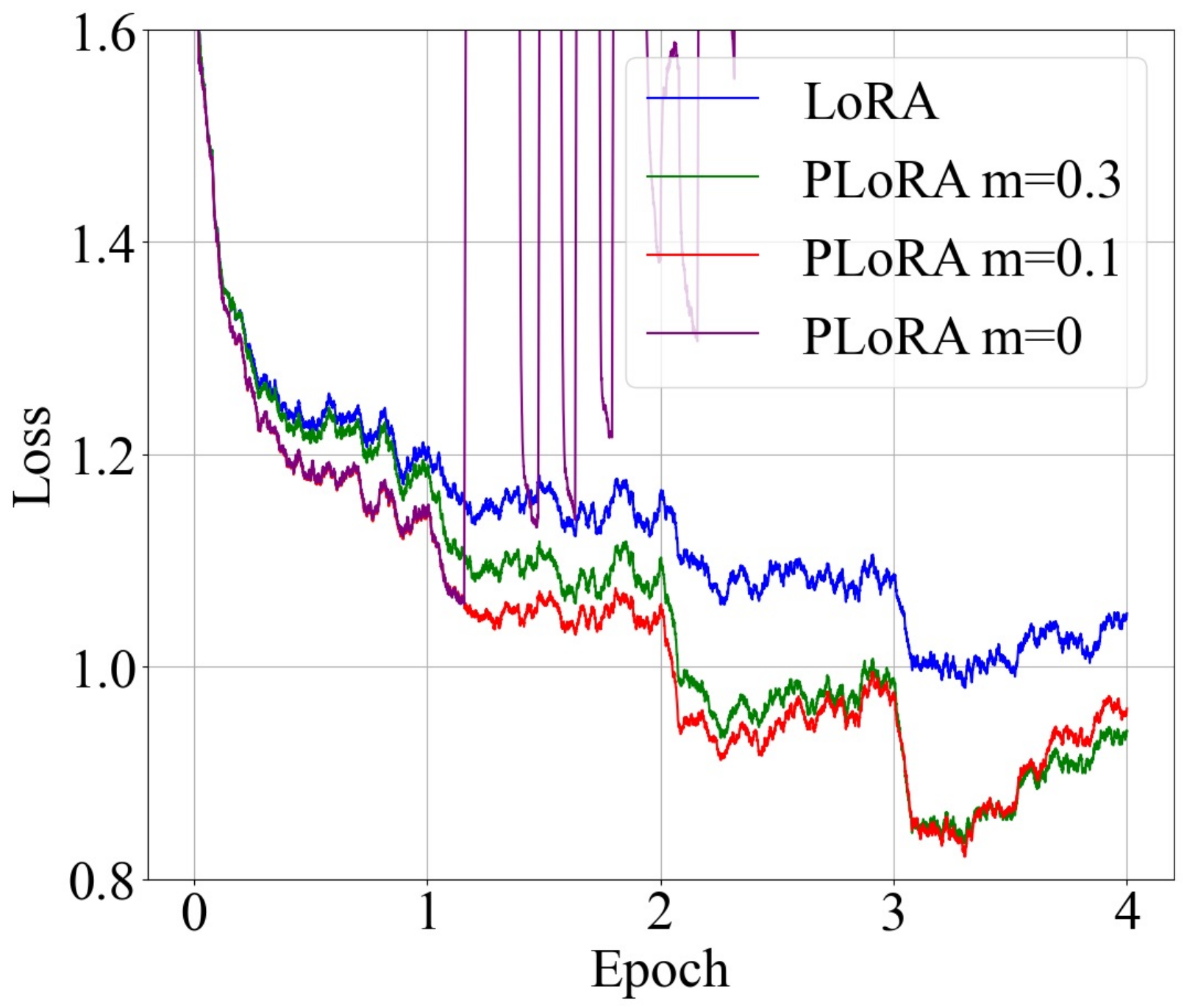}
  \end{center}
  \caption{Fixed rank to 8, train LoRA and PLoRA for 4 epochs. Among them, PLoRA includes three settings of m: 0, 0.1, and 0.3. To facilitate the observation of loss changes, we applied a smoothing window of 150 to the image. }
  \label{fig:loss_momentum}
\end{figure}

\begin{figure*}[!t]
  \begin{center}
    \includegraphics[width=0.98\textwidth]{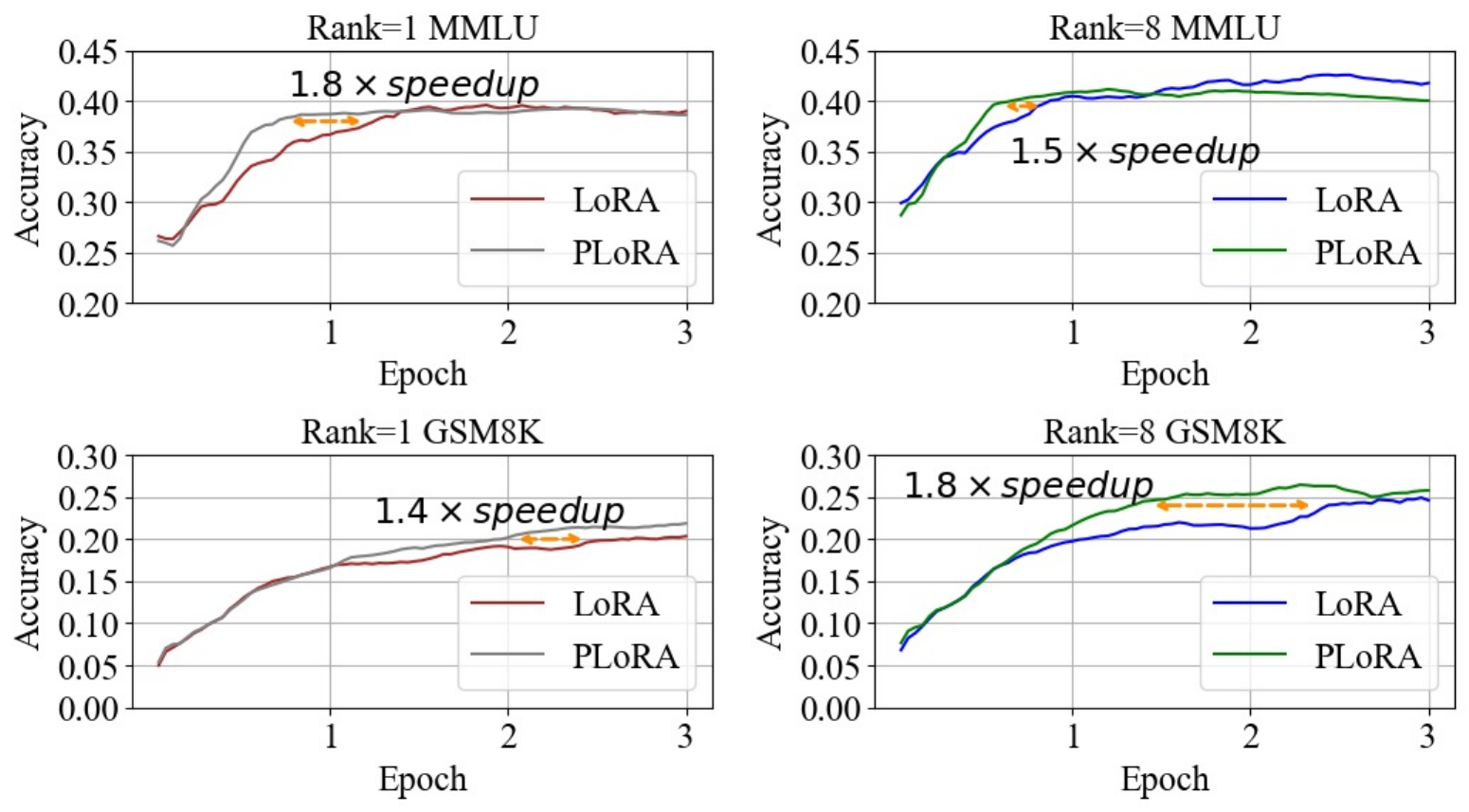}
  \end{center}
  \caption{We performed a detailed evaluation of the training process for four models with rank of 1 and 8 using two methods, LoRA and PLoRA. The models were saved every 100 steps, and checkpoints were evaluated using two metrics, GSM8K and MMLU, to obtain benchmark indicators for the models under different training settings. It is important to mention that a sliding window with a size of 10 was utilized for smoothing in order to enhance observation clarity.}
  \label{fig:capacity}
\end{figure*}

To investigate whether increasing the rank-up on mini-batch can surpass LoRA's learning ability limit, we conducted exploratory experiments on unloading points using a rank of 1. As shown in Table \ref{table:unloading}, solely increasing the amount of training data will not only enhance this stage of LoRA training, but also reduce the frequency of unloading within one epoch, leading to decreased training effectiveness. 
Based on empirical evidence, we set 4.8k data as the unloading point for PLoRA during training.

\subsection{Momentum Setting}

In our PLoRA experiments, we incorporated momentum to adjust unloading. Rather than completely replacing all $A$ and $B$ matrices as in the standard approach, we selectively update LoRA, scaling the $BA$ product by $(1-m)$ and $A$ and $B$ matrices by $m$. This method addresses the issue of inconsistent loss across mini-batches due to dataset variability, which is especially pronounced in complex tasks like GSM8K. Fully resetting matrices per mini-batch can degrade training and risk model failure.
Momentum improves training stability by not fully trusting each mini-batch result, providing a consistent update direction before training the subsequent batch. This approach, building on a predetermined unload point, allows for uninterrupted training and aids in pushing LoRA training boundaries.
 
As demonstrated in Figure \ref{fig:loss_momentum}, when the momentum of PLoRA is set to zero, the loss becomes chaotic after completing the first epoch and does not recover with further training. This suggests that PLoRA with a momentum of 0 collapses after the second epoch under a rank of 8. However, when the momentum is set to 0.1 or 0.3, the training with PLoRA remains stable. Additionally, we noticed that lower values of momentum result in smaller losses during training while maintaining stable training conditions. When increasing the number of epochs in training, we observed that higher momentum in PLoRA tends to have lower training losses compared to those with lower momentum towards the end approximately fourth epoch's completion time frame. Taking into account observations such as experiencing lower loss but encountering training collapse without any momentum in the first epoch; we can infer that lower momentum enable faster fitting to training data while higher momentum ensure stability during longer periods of training and lead to better fit overall.

 \begin{figure}[t]
  \begin{center}
    \includegraphics[width=0.45\textwidth]{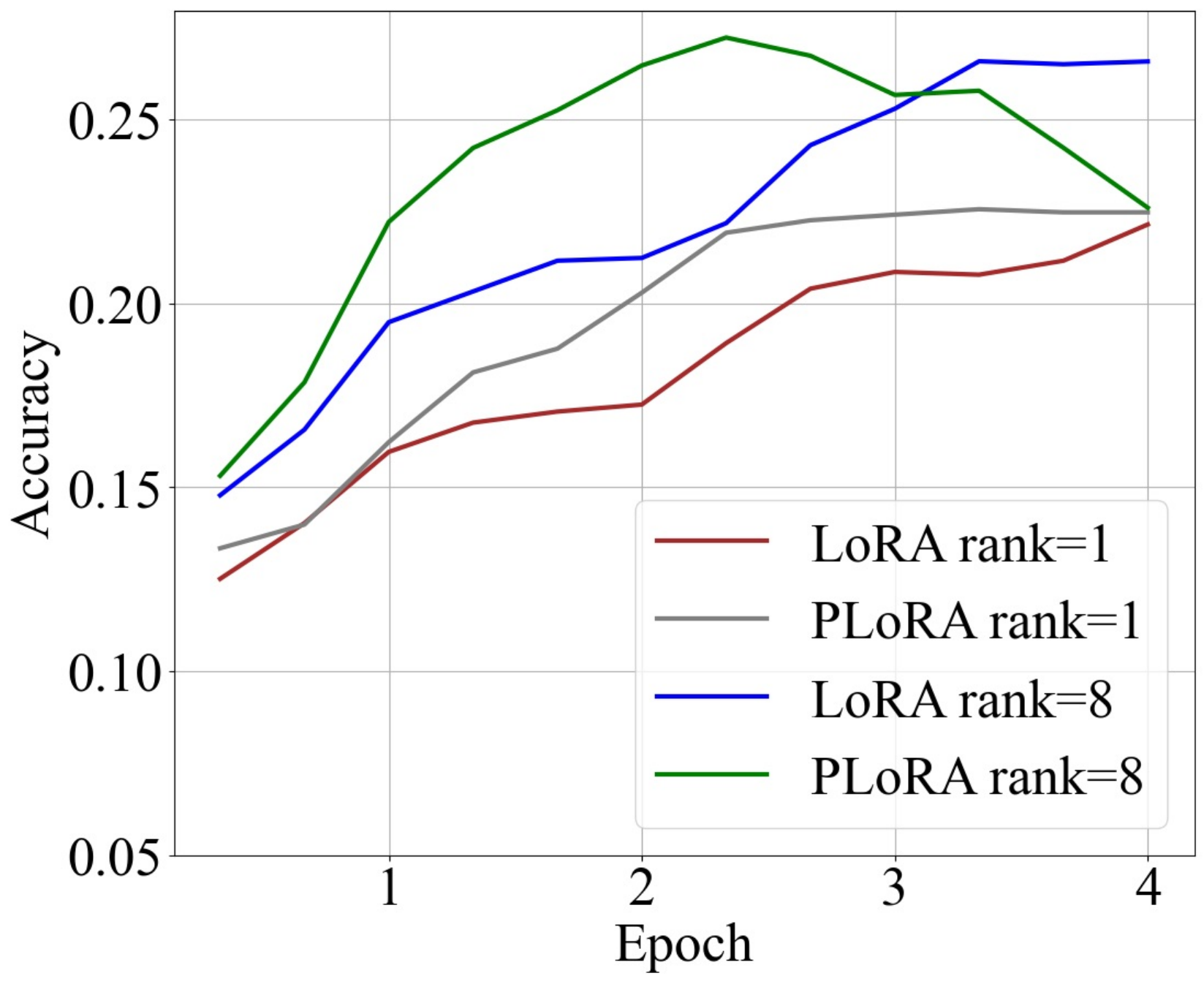}
  \end{center}
  \caption{To analyze the changes in checkpoint indicators during GSM8K evaluation, we continue training the models of the four settings for 4 epochs. Unlike Figure \ref{fig:capacity}, our focus here is on observing curves' changes over a longer training period. Therefore, we will save a checkpoint every 800 steps.  }
  \label{fig:capacity_long}
\end{figure}

\subsection{Training Stages}

\begin{table*}[ht]
    \centering
    \resizebox{0.85\textwidth}{!}{
    \begin{tabular}{c|ccccc|cc}
    \toprule
        & \multicolumn{5}{c|}{\textbf{LoRA}} & \multicolumn{2}{c}{\textbf{PLoRA}} \\
        \midrule
        Updating Modules & $W_q, W_v$ & $W_q, W_v$ & $W_q, W_v$ & $W_q, W_v$ & {all $W_*$} & {all $W_*$} & {all $W_*$} \\
        Rank & 1 & 1 & 1 & 8 & 1 & 8 & 8  \\
        Trainable Params &0.5M&0.5M&0.5M&4.2M&2.5M&20.0M&20.0M \\
        \midrule
        Learning Rate&2e-5&1e-4&5e-4&1e-4&1e-4&1e-4&5e-4 \\
        \midrule
        GSM8K&7.9&11.8&14.0&14.7&16.3&24.0&17.3 \\
        \bottomrule
    \end{tabular}}
    \caption{Apply LoRA to various weights of the model, utilize different learning rates, and evaluate on the GSM8K benchmark after training for one epoch. In the figure, all $W_*$ denotes the application of LoRA to all seven linear layers, including self-attention and multilayer perceptron. It can be observed that by applying LoRA to all linear layers, the model with a rank of one (trained with fewer trainable parameters) achieves better performance on GSM8K compared to the model with a rank of eight where LoRA is only applied to $W_q$ and $W_v$.}
    \label{table:settings}
\end{table*}

We conduct more detailed experiments to further demonstrate PLoRA's capabilities at different training stages. Through additional analysis, we also gain insights into unsatisfactory results and further validate that PLoRA outperforms LoRA.

Particularly, we extend the length of training and save checkpoints every 100 training steps. 
The unloading point of PLoRA is set at 400 steps.
The saved checkpoints are evaluated on two commonly used benchmarks: GSM8K and MMLU. This experimental setup allows deeper examination of the models' performance at various training stages.

Figure \ref{fig:capacity} shows that PLoRA consistently outperforms the ordinary LoRA method in the early stages of training, particularly on GSM8K. However, as training progresses, there is a gradual decline in the PLoRA curves and an increase in the LoRA metrics. 
The orange dashed line in Figure \ref{fig:capacity} further shows that PLoRA method exhibits significant acceleration during the early and middle stages of training. With equal computational resources and training speed, PLoRA can save about half of the training time compared to LoRA.

PLoRA also tends to overfit from the third epoch onward, as evidenced in both rank configurations, with MMLU effects being more severe. As depicted in Figure \ref{fig:capacity_long}, PLoRA achieves considerably lower loss than LoRA at epoch three for rank 8 across four epochs, yet LoRA excels in MMLU performance. Additionally, the figure shows a marked decline in PLoRA's performance when training is extended to six epochs.

We attribute the overfitting of PLoRA to a similar phenomenon in full fine-tuning. LoRA mitigates overfitting through low-rank $\Delta W$ updates, but PLoRA's accumulation of low-rank matrices ultimately results in a high-rank $\Delta W$, akin to full fine-tuning, leading to overfitting. 
PLoRA absorbs more knowledge from the training set than LoRA, compromising its generalization and causing overfitting, aligning with our theory that PLoRA overcomes the learning limitations imposed by the rank of $\Delta W$ in mini-batch LoRA training.

As per Figure \ref{fig:capacity_long}, after four epochs, the rank 1 model's performance on GSM8K declines, while the rank 8 model sees this decline in the third epoch. GSM8K peaks later than MMLU, implying that GSM8K is more challenging and demands higher learning capabilities. Table \ref{table:results} shows PLoRA's superior performance on GSM8K, indicating its suitability for difficult tasks requiring significant learning abilities. The earlier overfitting in rank 8 models compared to rank 1 also points to greater learning capabilities for higher ranks in the same task.

\subsection{Learning rate and PLoRA Layers}
\label{sec:LoRAs_Settings}
We delve deeper into the two most critical parameters for implementing PLoRA: the learning rate and the number of linear layers applied with LoRA.

In the current large language model training, maximizing the improvement of LoRA training ability is a focus of using the LoRA method. We have also summarized empirical experimental results in the process of experimentation. From Table \ref{table:settings}, it can be seen that when only changing the learning rate and keeping other settings fixed, a larger learning rate achieves better results after 1 epoch of training. Unlike full fine-tuning large language models using a universal learning rate of 2e-5, in LoRA training, it actually needs to start training from the initialized LoRA matrix $A B$, so a higher learning rate is more suitable for LoRA training. However, since PLoRA's method accumulates LoRA training on mini-batches, if too high of a learning rate is used, it will instead reduce the effectiveness of learning. From Table \ref{table:settings}, we can observe that when using PLoRA to train the model with a rank of 8, setting a learning rate of 5e-4 has much lower performance compared to setting a learning rate of 1e-4.

In terms of the number of linear layers applied with LoRA, we mainly focus on a applying LoRA to all linear layers.
The motivation of applying LoRA to all linear layers is that to compare it with full fine-tuning. The high-rank updates obtained only by using the PLoRA method on $W_q W_v$ are also applied to only a small part of the model. Applying LoRA to all linear layers is also a necessary step for the cumulative high-rank updates of the PLoRA method to reach the theoretical upper limit of full fine-tuning. In Table \ref{table:settings}, we also found that models applying LoRA to all linear layers can achieve better training results with fewer training parameters compared to models that only apply it to $W_q W_v$ but have higher rank.

\section{Conclusion}
We propose PeriodicLoRA (PLoRA), which is a Parameter-Efficient Fine-tuning method based on the low-rank adaptation theory. We assume that the biggest gap between full fine-tuning and LoRA lies in the fact that the update matrix of LoRA is much lower than that of full fine-tuning. By using accumulated LoRA weights on mini-batches, we obtain higher-rank update matrices during training compared to regular LoRA, thereby improving the training effectiveness of LoRA without increasing GPU usage. Our method is not complicated and theoretically can achieve the same training effectiveness as full fine-tuning for LoRA fine-tuning. To better utilize PLoRA, we propose a complete model training approach and analyze experimental data to demonstrate that high-rank weight updates obtained from accumulating low-rank weights have stronger learning capabilities than long-term trained low-rank weights.
\section{Limitations}
Although PLoRA has achieved exciting results, there are some limitations in our current research that are worth acknowledging. This paper mainly evaluates the effectiveness of PLoRA in the scenario of instruction fine-tuning for difficult tasks. However, in multimodal tasks with more complex datasets involving image and text, it is still worth studying how our PLoRA can be better applied to these tasks. Our PLoRA method provides insights into using LoRA to fit the training process of full fine-tuning, but providing a strict explanation for this process remains challenging.



\bibliography{custom,anthology}



\end{document}